\newif\ifarXiv
\arXivtrue 

\documentclass[3p]{elsarticle}
\usepackage{graphicx} 
\usepackage{xcolor}
\usepackage{rotating}
\usepackage{tabularx}
\usepackage{booktabs}
\usepackage{url}
\usepackage{multirow}
\newcommand{\red}[1]{\textcolor{red}{#1}}

\newcommand{\blue}[1]{\textcolor{blue}{#1}}
\usepackage{amsmath}
\usepackage{float}
\usepackage{comment}
\usepackage{multirow}

\journal{Theoretical Computer Science}

\begin{document}
\begin{frontmatter}

\title{BIPOLAR: Polarization-based granular framework for LLM bias evaluation}
\author[i1]{Tom\'a\v s Filip}
\ead{tomas.filip@osu.cz}
\author[i1]{Martin Pavl\'\i\v cek}
\ead{martin.pavlicek@osu.cz}
\author[i1,i2]{Petr Sos{\'\i}k\corref{cor1}}
\ead{petr.sosik@osu.cz}
\cortext[cor1]{Corresponding author}

\affiliation[i1]{organization={Institute for Research and Applications of Fuzzy Modeling, University of Ostrava},
            addressline={30. dubna 22}, 
            city={Ostrava},
            postcode={70200}, 
            country={Czech Republic}}

\affiliation[i2]{organization={Institute of Computer Science, Faculty of Philosophy and Science, Silesian University in Opava},
            addressline={Bezručovo náměstí 1150/13}, 
            city={Opava},
            postcode={74601}, 
            country={Czech Republic}}

\begin{abstract}
Large language models (LLMs) are known to exhibit biases in downstream tasks, especially when dealing with sensitive topics such as political discourse, gender identity, ethnic relations, or national stereotypes. Although significant progress has been made in bias detection and mitigation techniques, certain challenges remain underexplored. This study proposes a reusable, granular, and topic-agnostic framework to evaluate polarisation-related biases in LLM (both open-source and closed-source). Our approach combines polarisation-sensitive sentiment metrics with a synthetically generated balanced dataset of conflict-related statements, using a predefined set of semantic categories.

As a case study, we created a synthetic dataset that focusses on the Russia-Ukraine war, and we evaluated the bias in several LLMs: Llama-3, Mistral, GPT-4, Claude 3.5, and Gemini 1.0.
Beyond aggregate bias scores, with a general trend for more positive sentiment toward Ukraine, the framework allowed fine-grained analysis with considerable variation between semantic categories, uncovering divergent behavioural patterns among models. Adaptation to prompt modifications showed further bias towards preconceived language and citizenship modification.

Overall, the framework supports automated dataset generation and fine-grained bias assessment, is applicable to a variety of polarisation-driven scenarios and topics, and is orthogonal to many other bias-evaluation strategies. 
\end{abstract}


\begin{highlights}
\item Novel granular framework for evaluating LLM bias on polarising themes.
\item Orthogonal to standard bias metrics and measures.
\item Semi-automated generation of synthetic theme-specific datasets. 
\item Case study on the Russia-Ukraine conflict.
\end{highlights}


\end{frontmatter}

\section{Introduction}
\label{sec_intro}

Large language models (LLMs) have become foundational in modern natural language processing (NLP), powering a wide range of applications from machine translation to sentiment analysis. However, their susceptibility to bias, particularly in politically\cite{linegar2023large}, nationally, and socially sensitive contexts\cite{10.1145/3442188.3445924}, poses significant challenges for their safe and equitable deployment \cite{blodgett2020language, feng2023pretraining, chu2024fairness}, especially when downstream tasks are related to LLM output \cite{stureborg2024large, nogara2024toxicbiasperspectiveapi, Ferrara_2023}. Understanding and mitigating such biases remains a critical concern in both academic research and real-world applications.

In this paper, we present a scalable and reusable BIPOLAR framework for detecting and analysing sentiment bias in LLMs, tailored to polarised and conflict-driven domains. This work relates to our previous research on LLM-based sentiment classification in polarity-based statements \cite{filip2024}, responding to the general need for systematic and interpretable methods to uncover latent political or ideological tendencies in model behaviour where current approaches to bias detection fall short or cannot be applied.

The first step is the (semi-)automated construction of a synthetic, polarity-balanced dataset centred on conflict-related events, using a structured annotation scheme guided by a codebook or a nested ontology with a set of semantic categories. The dataset integrates multiple semantic dimensions, including sentiment polarity, entity role (e.g., subject or object of the statement), and event category. This level of control facilitates the direct attribution of biased behaviour to specific input properties, supporting more transparent model evaluations. The responses of the examined LLM then serve as inputs to a symmetry-based sentiment evaluation scheme, allowing for their fine-grained analysis.

As a case study, we generated a synthetic dataset centred on the Russia-Ukraine conflict, based on the categories of the CAMEO (Conflict and Mediation Event Observations) codebook\footnote{\url{http://data.gdeltproject.org/documentation/CAMEO.Manual.1.1b3.pdf}}. Despite the geopolitical importance of this domain, structured evaluations of LLM behaviour in this context remain limited \cite{urman2023silence, wadhwani2023sentiment}. We assess five LLMs: LLaMA 3, Mistral, GPT-4, Claude 3.5, and Gemini 1.0 on a set of symmetrical prompts with statements designed to elicit sentiment responses toward defined entities. Our approach enables direct, model-agnostic comparison through numerical scoring, clustering analysis, and visualisation.

\textbf{Main contributions:}
\begin{itemize}
\item We introduce a structured, step-by-step (semi)automated methodology for generating synthetic conflict-driven datasets with multilayered metadata, including sentiment polarity, entity roles, granular event categories, temporal frame, and entity description, derived from a chosen ontology.

\item We propose a symmetry-based sentiment evaluation scheme that facilitates interpretable, quantitative, and visual comparison of model responses.

\item We apply the methodology and a reduced synthetic dataset to a polarised domain of the Russia–Ukraine conflict and demonstrate its utility to identify bias patterns in multiple LLMs. We also outline its extensibility to other domains (e.g., US-–China, Israel-–Iran, China--Taiwan) and socially polarised topics (e.g., immigration--antimigration, vaccination--anti-vaxxers).
\end{itemize}

The remainder of this paper is structured as follows. Section~\ref{sec_background} reviews related work. Section~\ref{sec_methods} outlines our dataset construction and evaluation framework. Section~\ref{sec_experiments} presents the experimental results of our use case. Section~\ref{sec_cdiscussion} discusses the interpretation and limitation of the results together with their future extension, and Section~\ref{sec_conclusion} contains concluding notes.


\section{Background}
\label{sec_background}

\subsection{Bias Introduction}

Bias in large language models (LLMs) can reflect or reinforce harmful stereotypes, discrimination, and unfair treatment of individuals or groups. The bias often originates from multiple sources, including training data, data labelling procedures, and internal representations such as embeddings \cite{blodgett2020language, feng2023pretraining, liang2021towards,gallegos2024bias}. 

The training data used to pre-train LLMs is a primary source of bias, as it frequently contains historical prejudices and representation imbalances. These issues arise when certain groups, viewpoints, or contexts are systematically underrepresented or misrepresented \cite{feng2023pretraining, zhao2019gender, caliskan2017semantics}. For example, training corpora may perpetuate occupational and gender stereotypes, such as associating technical roles such as 'programmer' predominantly with males \cite{zhao2019gender}. 

Labelling practices can also introduce bias. Annotator demographics, subjective guidelines, and culturally skewed interpretations may influence the way text is labelled, leading to unintended consequences in downstream applications \cite{nadeem2020stereoset, glaese2022improving, shah2019predictive,gallegos2024bias}.

Model embeddings (dense vector representations of tokens) capture semantic regularities, but also tend to encode undesirable social associations. This can result in biased clustering (e.g., associating certain professions with specific genders or ethnicities), which then propagates into task performance and decision-making in applications such as classification or summarization \cite{may2019measuring, sweeney2019transparent}.

\subsection{Bias Evaluation}

Bias in LLMs can be systematically assessed using a combination of \textbf{embedding-based}, \textbf{probability-based}, and \textbf{generated text-based} metrics. Each provides a different analytical lens to examine different components of model behaviour \cite{chu2024fairness, bai2024measuring, lin2024investigating,li2024surveyfairnesslargelanguage}.

\textbf{Embedding-based metrics} evaluate geometric relationships in the latent space of the model. Techniques such as the \textit{Word Embedding Association Test} (WEAT) \cite{caliskan2017semantics} and \textit{Sentence Encoder Association Test} (SEAT) \cite{may2019measuring} compare cosine similarities between target and attribute vectors. For example, SEAT can reveal whether profession terms like “soldier” or “nurse” are more closely aligned with gender concepts, indicating encoded stereotypes.

\textbf{Probability-based metrics} assess the conditional likelihoods assigned to alternative continuations of controlled prompts. These include log-probability comparisons, masked language modelling evaluations, and benchmark datasets like \textit{StereoSet} \cite{nadeem2020stereoset} and \textit{CrowS-Pairs} \cite{nangia2020crowspairschallengedatasetmeasuring, liang2021towards}, which measure whether models systematically favour biased or stereotypical completions.

\textbf{Generated text-based metrics} analyse the sentiment, toxicity, or stance of freely generated outputs. This can be done using sentiment analysers \cite{nasukawa2003sentiment}, toxicity classifiers (e.g., Perspective API), or stance detection models \cite{hamborg2021newsmtsc}. For example, prompting a model with neutral descriptions of social or political entities and then analysing its outputs can reveal patterns of implicit bias \cite{perez2022discovering, dhamala2021bold,gallegos2024bias}.

Together, these metrics form a comprehensive suite for LLM bias evaluation. They enable researchers to assess not just what a model says, but also what it encodes, prefers, or avoids, thus revealing the multifaceted nature of bias in modern NLP systems. 

\section{Methods}
\label{sec_methods}
\subsection{Synthetic dataset}

We outline the methodology of building a structured synthetic dataset focused on polarising geopolitical topics. The dataset design combines insights from previous research \cite{filip2024} on social network discourse (e.g., X, YouTube, Reddit, Telegram) and employs a conflict-event ontology based on a chosen set of structured semantic categories. The construction process consists of several stages designed to ensure balance, symmetry, and robustness:

\begin{enumerate}
    \item \textbf{Polarized topic}: A publicly perceived polarising or conflict-related topic is selected, defining the two main target entities that are compared.
    
    \item \textbf{Categories and subcategories}: Suitable event categories and subcategories based on the chosen ontology are identified, such as \emph{fight}, \emph{assault}, \emph{cooperate}, etc. Categories can be nested and act as topic domains \cite{10.1145/3442188.3445924}. 
    
    \item \textbf{Events}: Representative events for each category are chosen. Events are constructed in pairs of opposing semantic terms (for example, 'calls for peace' vs. 'calls for war') to maintain polarity symmetry. 
    
    \item \textbf{Entities}: The target entities are selected in symmetric pairs. These may include countries (Russia, Ukraine), governments, political leaders (e.g., Putin, Zelenskyy), militaries, or sub-entities such as cities, weapon systems, or any other entity that can be constructed with a mirror approach.
    Entities can be described with granular metadata.
    
    \item \textbf{Entity role:} The entity can be a subject or object of the statement (\emph{Russia attacked}, \emph{Russia was attacked}).
    
    \item \textbf{Polarity:} A mirror polarity pair -- positive or negative (\emph{Russia did not attack}, \emph \textbf{Russia  attacked}).

    \item \textbf{Temporal frame:} The statement may occur in the past, present, future (\emph{Russia attacked, Russia is attacking, Russia will attack}).
\end{enumerate}

From each combination of topic, category, event, entity, role, polarity, and temporal frame, a final statement is generated and used as a sample in the dataset. Whenever a topic and an ontology are chosen, the remaining tasks could be automated with the help of an LLM.
This results in a dataset where every entity is evaluated across all event categories, with balanced sample counts for each entity and event polarity. Importantly, while we use "positive" and "negative" labels to organise the statements, these are not revealed to the evaluated models. 

\subsection{Prompt variants for bias  shift analysis}

It is known that LLMs often adapt their responses according to the nationality, location, language etc. of the interviewer. Inspired by Antrophic's linguistic and cross-national prompts \cite{durmus2023towards} and general sensitivity to prompts\cite{hida2024social}, we propose to investigate LLM bias under different identity, language, and contextual framing, and we include in our framework several prompting strategies:

\begin{itemize}
\setlength\itemsep{-3pt}
    \item Statement written in the language of Entity 1 (e.g., Russia)
    \item Statement written in the language of Entity 2 (e.g., Ukraine)
    \item Complete prompts in the language of Entity 1
    \item Complete prompts in the language of Entity 2
    \item English prompts that suggest the perspective of a citizen of Entity 1
    \item English prompts that suggest the perspective of a citizen of Entity 2
    \item Prompts with explicit real-world contextual information about current events
\end{itemize}

Employing this strategy of prompt definition (language, citizen description, in-context knowledge) enables one to measure a shift in the sentiment response due to latent LLM ``national" biases.  

\subsection{Evaluation framework}
\label{sec:eval_framework}

The BIPOLAR framework employs generation-based metrics inspired by the works cited in the previous paragraph. We measure sentiment-based bias using controlled prompts generated from the dataset. Each prompt consists of a single statement = sample from the dataset, together with a common preamble that can adopt the form, for example:

\begin{quote}
\emph{For the following text, return the sentiment of the entire text. Return the sentiment score of the event, where 0 is negative sentiment and 100 is positive sentiment. Do not append any justification, just return the score.}
\end{quote}

By collecting sentiment scores across all models, events, and entities, we transform the abstract problem of bias detection into \emph{quantitative comparison of sentiment distributions}. The crucial methodological element is the \textbf{symmetry} of the statements, which allows one to detect systematic bias as a divergence in the average sentiment towards opposing entities. 
So, for the bias evaluation, it is not the individual sentiment scores that are important, but their comparison on the level of different models, statement polarities (positive/negative), statement categories, prompt variants, etc.
Our sample dataset and the evaluation code are available as open resources; see Section ``Data availability''.


\subsection{Models}

We used the following language models without any fine-tuning or few-shot learning:

\medskip\noindent
    \begin{tabular}{lll}
    \hline
      \bf Model & \bf Params & \bf Web page\\
    \hline
 Llama-3-8B-Instruct & 8B & {\small https://huggingface.co/meta-llama/Meta-Llama-3-8B-Instruct}\\
 Mistral-7B-Instruct-v0.3 & 7.25B & {\small https://huggingface.co/mistralai/Mistral-7B-Instruct-v0.3}\\
 Claude-3-5-sonnet-20240620 & $\approx$ 175B & {\small https://claude.ai}\\
 Gemini-1.0-pro-002 & n/a & {\small https://gemini.com}\\
 GPT-4-0613 & n/a & {\small https://platform.openai.com/docs/models/gpt-4-and-gpt-4-turbo}\\
    \hline
    \end{tabular}


\section{Results}
\label{sec_experiments}

In this section, we present a case study based on a synthetic dataset prepared according to the guidelines described above. We chose 15 CAMEO-based semantic event categories, each with positive and negative variants, resulting in 30 subcategories. The study was restricted to using the present temporal frame and the subject entity role. Each event had a mirrored form for both Russia and Ukraine. We used default prompts, in-context prompts, statements in English, Russian, and Ukrainian, and citizenship specification for bias-shift detection. 

 If a model failed to produce a valid output, we repeated the query multiple times. Malformed outputs were excluded from the final evaluation. Interestingly, the models mostly assigned scores to events in multiples of 5, although they were not instructed to do so. The resulting sentiment scores are aggregated by categories and models and visualised in several ways targeting different facets of the bias. Detailed results containing sentiment scores assigned by the models to all the presented events are available online; see the section ``Data availability''.

\subsection{Radar plots of bias} \label{sec-radar}

Radar plots are an illustrative way of comparison of sentiment scores towards RU and UA targets over all categories, where the difference between the targets represents the bias of the model. Each figure contains plots targeting both RU (red) and UA (blue) for each category and for both positive and negative events. 

\begin{figure}
    \centering
    \includegraphics[width=.9\linewidth]{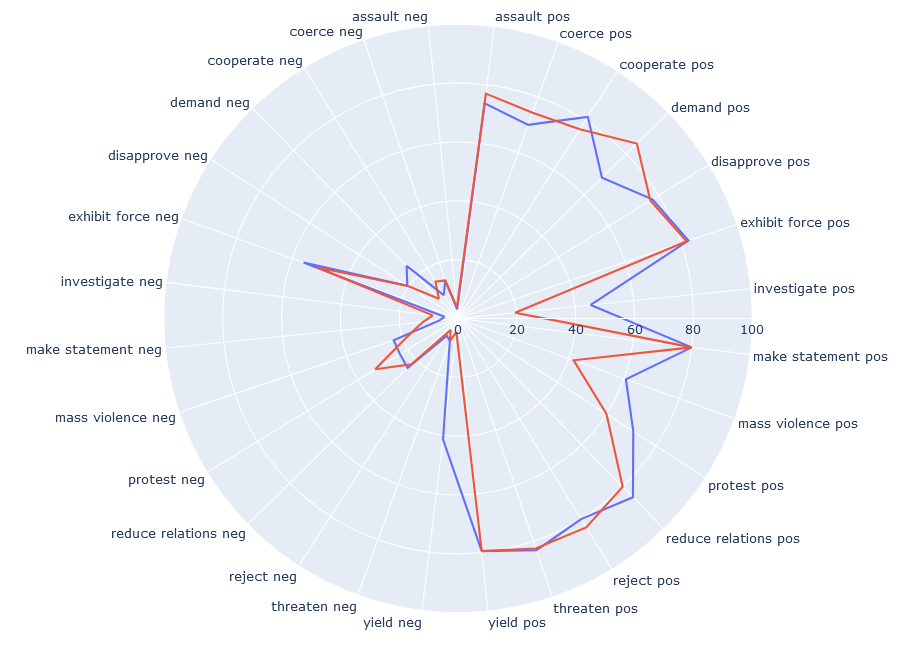}
    \includegraphics[width=.91\linewidth]{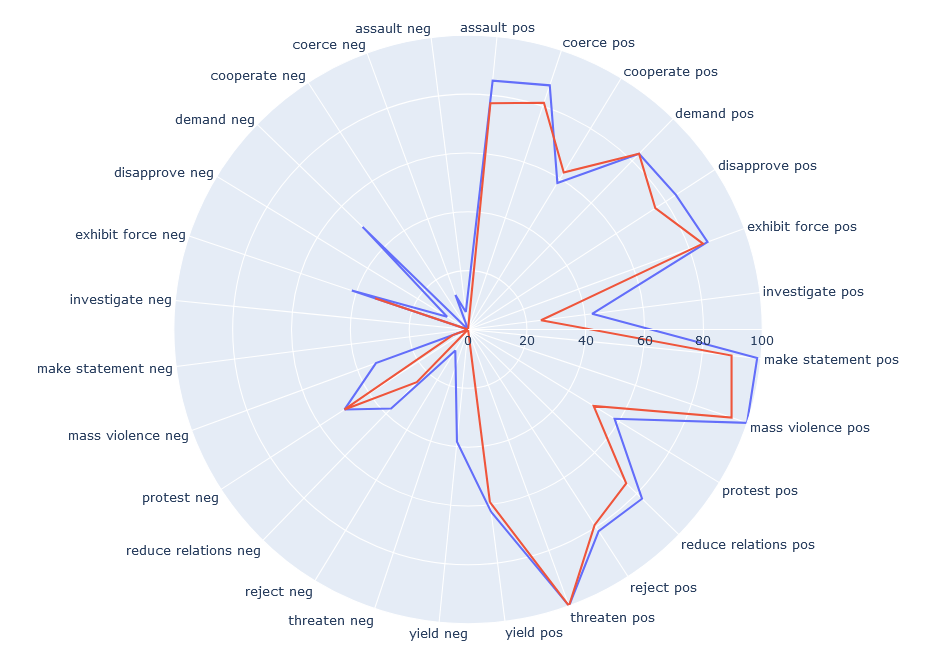}
    \caption{Mean sentiment scores per category, \red{RU}--\blue{UA}, models Llama (top) and Mistral (bottom)}
    \label{fig-radar-1}
\end{figure}

Llama 3 (Fig. \ref{fig-radar-1} top) exhibited the lowest bias among all models tested, especially for positive events. For negative events, the model was slightly biased towards Ukraine. 

Mistral (Fig. \ref{fig-radar-1} bottom) showed the highest differences between the scores for the positive and negative categories among all models, so its sentiment stances can be termed radical. Overall, the model had only a slight bias towards Ukraine for positive events, while the bias was higher for negative events. 

\begin{figure}
    \centering
    \includegraphics[width=.88\linewidth]{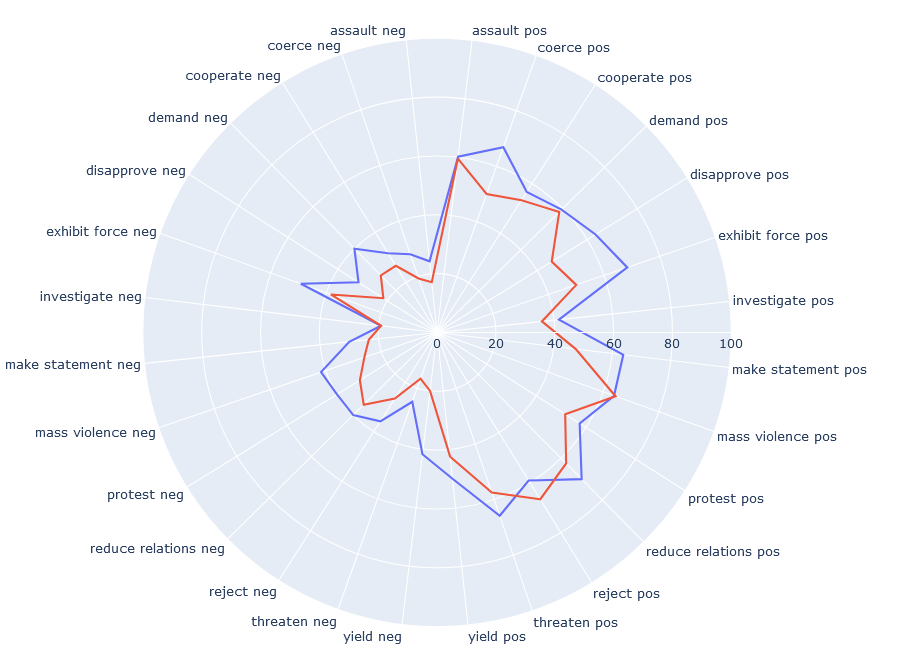}
    \includegraphics[width=.88\linewidth]{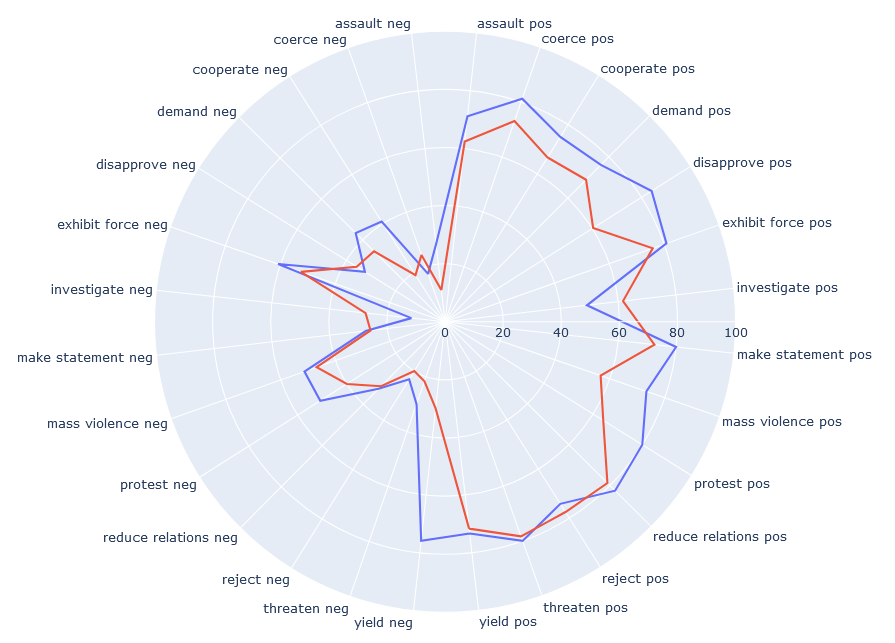}
    \caption{Mean sentiment scores per category, \red{RU}--\blue{UA}, models Claude (top) and Gemini (bottom)}
    \label{fig-radar-2}
\end{figure}

Claude (Fig. \ref{fig-radar-2} top) showed almost symmetric patterns for all categories. The sentiment score targeting Russia was lower for both positive and negative events, showing that the model was almost equally biased toward Ukraine in all categories, except \emph{fight} and \emph{reject} (both positive). 

Gemini (Fig. \ref{fig-radar-2} bottom) showed the highest bias towards Ukraine among all evaluated models in almost all categories, both positive and negative, and also the sentiment scores were higher overall.  

\begin{figure}[t!]
    \centering
    \includegraphics[width=.88\linewidth]{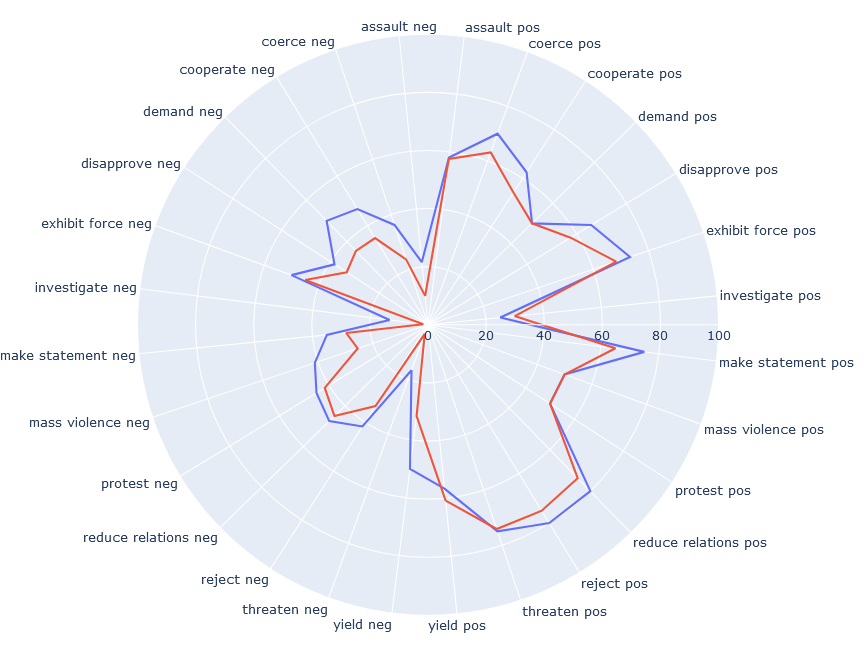}
    \caption{Mean sentiment scores per category, \red{RU}--\blue{UA}, model GPT4}
    \label{fig-radar-gpt}
\end{figure}

GPT4 (Fig. \ref{fig-radar-gpt}) generally showed a positive bias towards Ukraine similar to Claude, but significantly weaker for positive events. The bias remained higher for most of the negative events. The anomalous categories with favourable sentiment towards Russia were \textit{fight}, \textit{yield} and \textit{investigate.}

Generally, the results approximated the anticipated distribution of positive and negative sentiment. However, there was a visible anomaly in the category \textit{investigate} (positive event but negative sentiment in most models). The category \textit{protest} remains close to the neutral value, as well as the  category \textit{exhibit force} in a negative event.

\subsection{Divergence between RU and UA targets} \label{sec-entropy}

\begin{figure}[]
    \centering
    \includegraphics[width=.95\textheight,angle=-90]{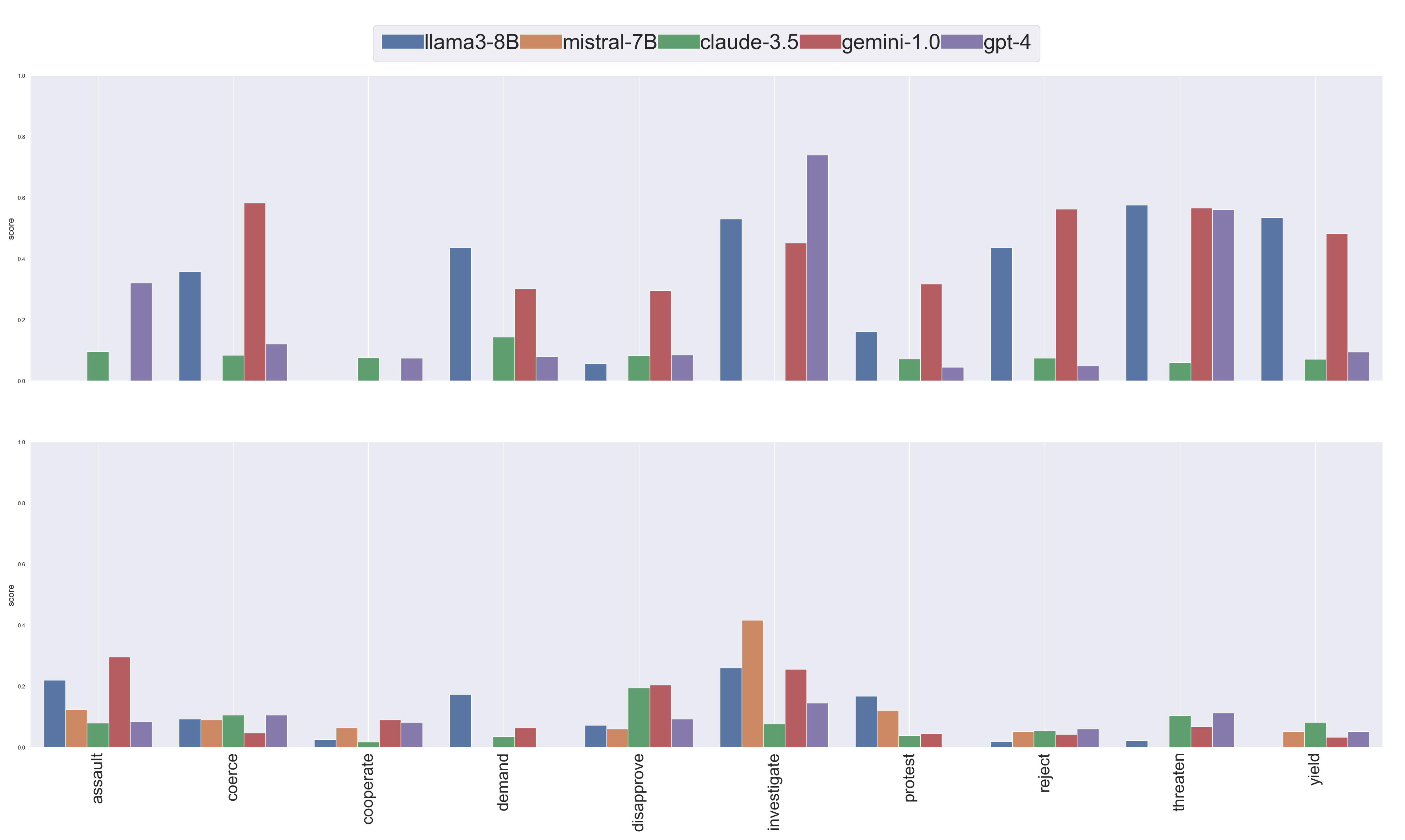}
    \caption{Ten categories with the highest Jensen-Shannon divergence between RU and UA sentiment score distributions. Left: positive statements; right: negative statements.}
    \label{fig_entropy}
\end{figure}

Figure \ref{fig_entropy} shows the Jensen-Shannon (JS) divergence that determines the similarity of the RU and UA sentiment distributions for a given category and model. The smaller the divergence value, the more similar the distributions are. The JS divergence uses the Kullback-Leibler (KL) divergence, but unlike it, the JS divergence is symmetric, i.e., for distributions P and Q, the JS divergence is computed as the weighted average of two KL divergences: 
\begin{eqnarray}
    JS(P\parallel Q) &=& (KL(P\parallel M) + KL(Q\parallel M)) / 2,\mbox{ where}\\
    M &=& {\frac {1}{2}}(P+Q).
\end{eqnarray}
 
The most divergent categories for each model (Figure \ref{fig_entropy}) were the following:
Llama 3 -- \textit{make statement}, \textit{yield}, \textit{investigate}, \textit{demand}, \textit{reject}, \textit{threaten} (all negative). Gemini -- \textit{yield}, \textit{investigate}, \textit{reject}, \textit{threaten} (all negative). Mistral -- \textit{investigate}, \textit{mass violence} (all negative). GPT4 -- \textit{investigate}, \textit{threaten}, \textit{assault} (all negative). Claude did not produce any strong divergences. 

\subsection{Cluster analysis} \label{sec-clusters}

\begin{figure}[]
    \centering
    \includegraphics[width=\linewidth]{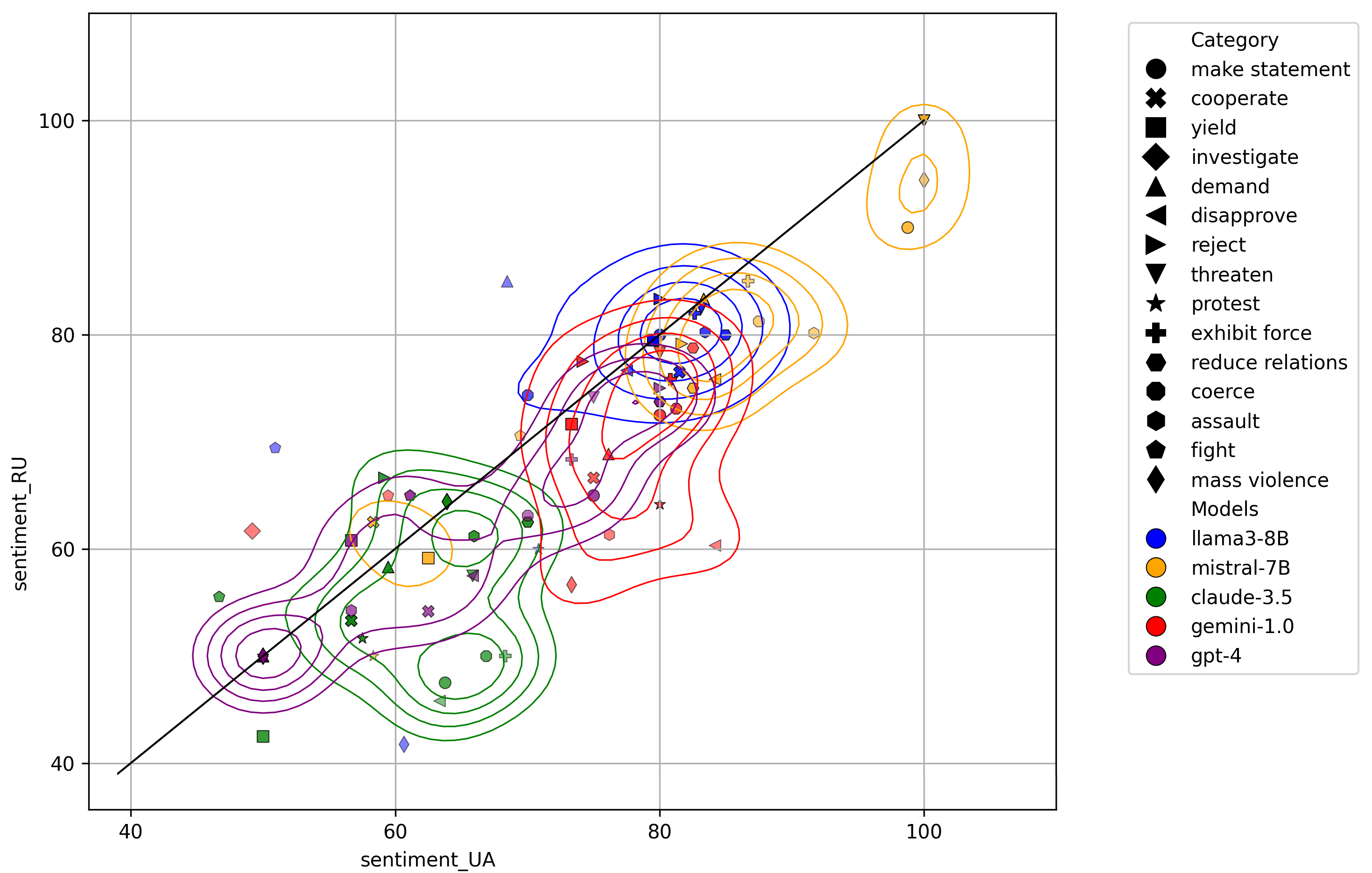}
    \includegraphics[width=\linewidth]{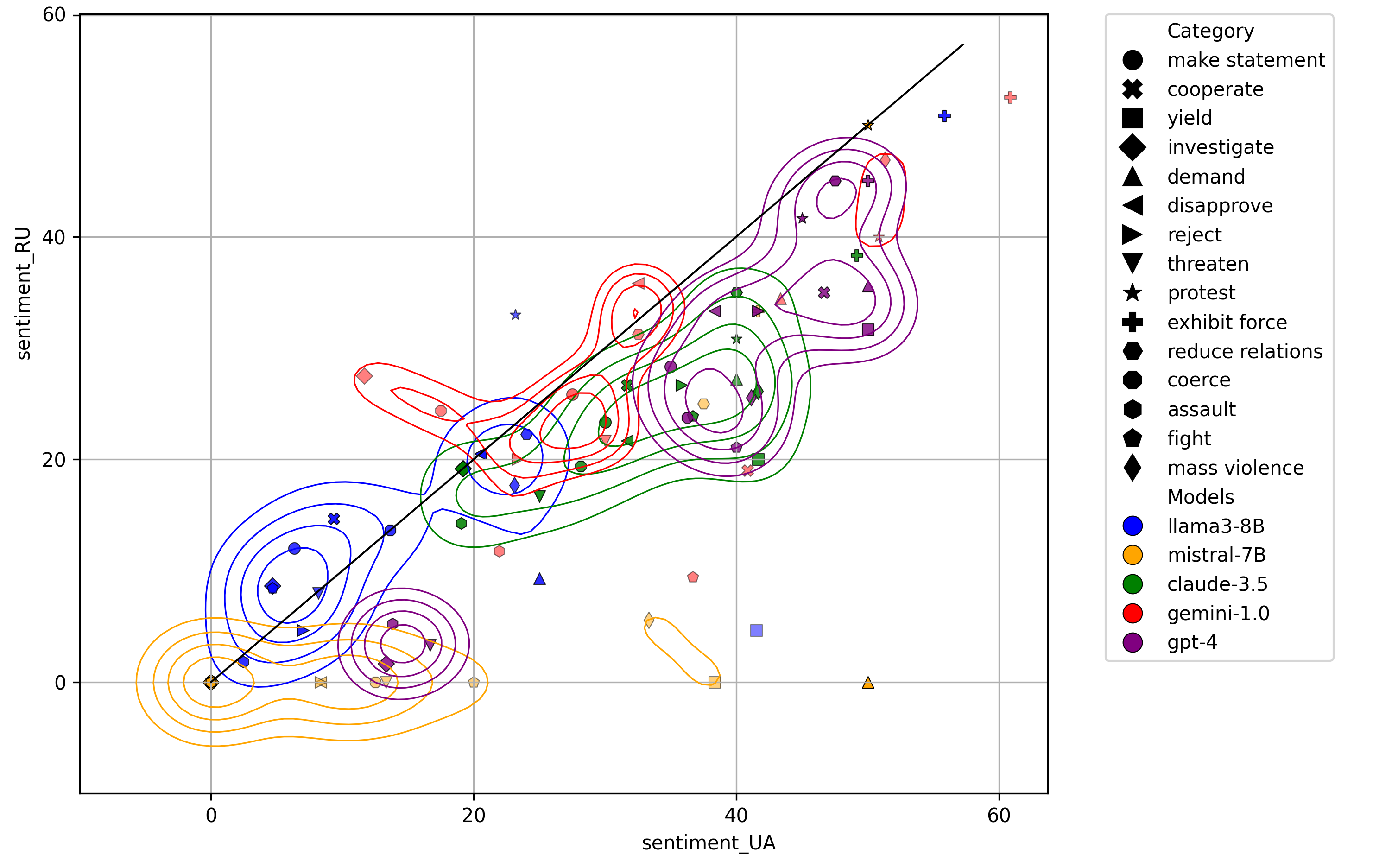}
    \caption{Scores of positive (above) and negative (bellow) events towards Russia and Ukraine, respectively. The bias is displayed as the deviation from the diagonal.}
    \label{fig-clustering-vanilla}
\end{figure}

Figure \ref{fig-clustering-vanilla} visualises the bias towards Russia versus Ukraine in the form of clusters corresponding to individual models. The coloured clusters represent the sentiment distributions of the individual models. The shapes of the data points correspond to categories of events. Each data point is the mean of all events in a given category. The X and Y axes correspond to sentiment scores towards Ukraine and Russia, respectively. Therefore, the bias is seen as the deviation of points from the diagonal (pro-Russian -- above the diagonal, pro-Ukrainian -- below the diagonal). This form of visualisation allows for each model to view its overall bias in comparison to other models, and, simultaneously, the bias distribution between categories.

\subsection{Bias distribution per categories} \label{sec-corellation}

\begin{figure}[]
    \centering
    \includegraphics[width=.8\linewidth]{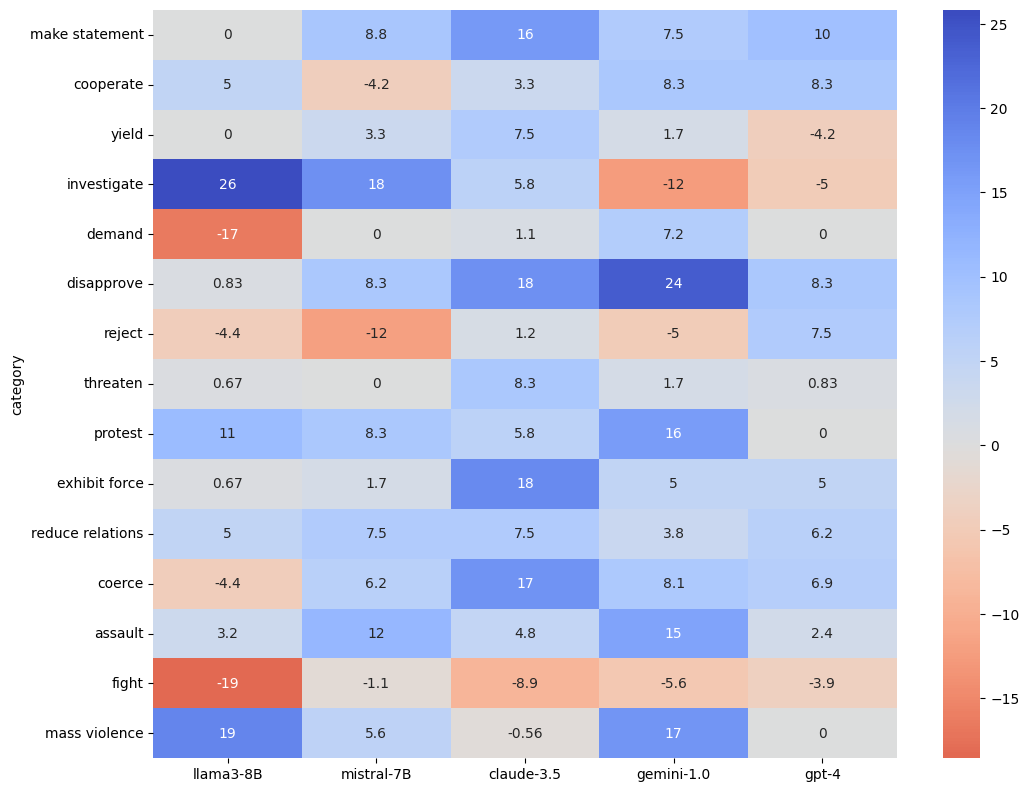}

    \includegraphics[width=.8\linewidth]{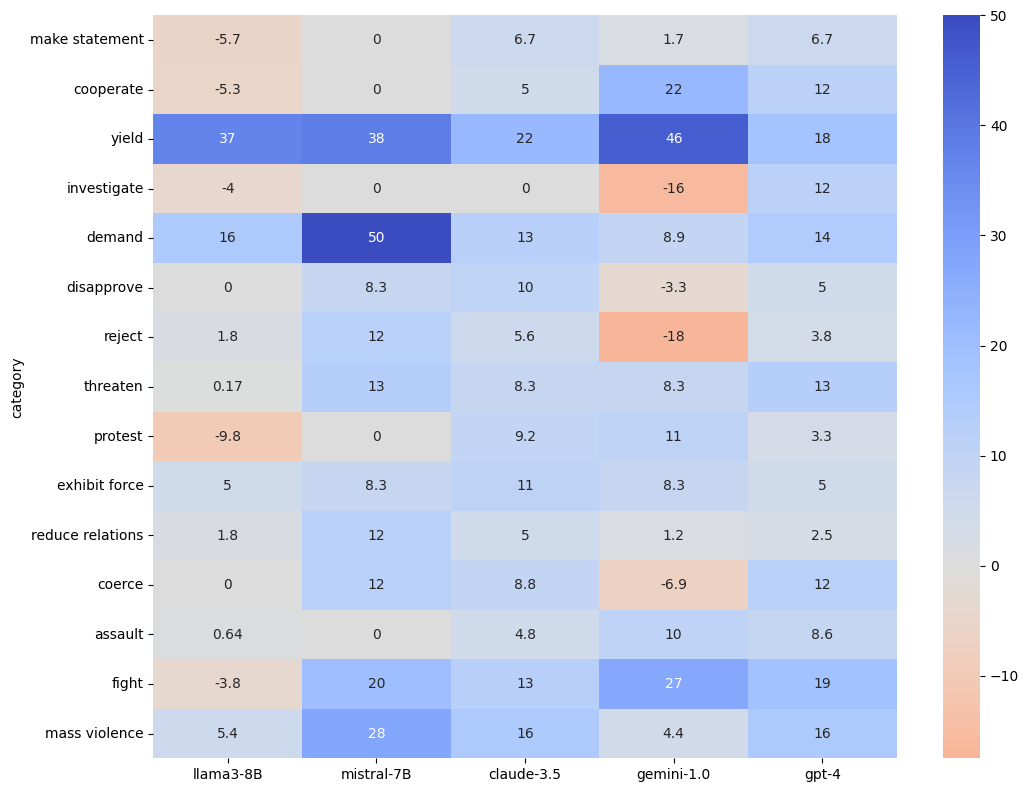}
    \caption{Average bias of sentiment (UA$-$RU) for \textit{positive} (above) and \textit{bellow} (bottom) statements.}
    \label{tab:heatmap}
\end{figure}

Figure \ref{tab:heatmap} visualises the bias distribution by categories and models. Blue tones denote bias towards UA, while red tones correspond to RU. Biases are calculated as differences between mean sentiments towards UA and RU per category and model. This form of visualisation is especially suitable for comparing the biases of individual categories.

\begin{equation}
    \mathrm{bias}(\mathrm{cat},\mathrm{model}) = \mathrm{mean\_sentiment}(\mathrm{UA}, \mathrm{cat},\mathrm{model}) - \mathrm{mean\_sentiment}(\mathrm{RU}, \mathrm{cat},\mathrm{model})
\end{equation}

\ifarXiv
A chord diagram visualising the largest differences in absolute bias between models for individual categories is provided in \ref{app_bias_diff}. 
\fi

To better understand the anatomy of bias of the tested models in individual categories, we defined two variables, \textit{bias consensus} and \textit{absolute bias} per category:

\begin{equation}\label{eq:consensus}
\mathrm{consensus}(\mathrm{cat}) = \underset{\mathrm{model}}{\mathrm{std\_dev}}(\mathrm{bias}(\mathrm{cat},\mathrm{model}))
\end{equation}

\begin{equation}\label{eq:abs_bias}
\mathrm{absolute\_ bias}(\mathrm{cat}) = \sum_\mathrm{model} |\mathrm{bias}(\mathrm{cat},\mathrm{model})|
\end{equation}

Both variables are calculated separately for positive and negative statements. Table \ref{tab:categories} compares the most/least biased categories with the most/least consensual ones. The clear trend is that the least biased categories coincide with the consensual ones, while the most bias-prone categories are controversial. This suggests that the models mostly do not share common bias trends towards certain categories. Instead, the most biased categories are those where the models' answers are imbalanced. There are a few exceptions with clear bias agreement across all models, such as the negative \textit{yield} category (biased towards UA) or the positive \textit{fight} category (bias towards RU).

\begin{table}
  \caption{Comparison of categories: (i) least/most biased by formula (\ref{eq:abs_bias}), (ii) consensual/controversial by formula (\ref{eq:consensus}).}
    \label{tab:categories}
    \centering
    \medskip
    \begin{tabular}{|l||l|l||l|l|}
\hline
   & \bf Least biased   &\bf Consensual 
& \bf Most biased & \bf Controversial\\
\hline
Positive  & threaten  &reduce rel. 
& investigate & investigate \\
statements& yield     &threaten    
& disapprove  & mass violence\\
          & demand    &yield       
& coerce      & disapprove\\
\hline
Negative  & make stat.  &exhibit force 
& yield  & demand \\
statements& reduce rel. &assault       
& demand & yield\\
          & assault     &reduce rel.   & fight  & fight\\
\hline
    \end{tabular}
  
\end{table}

\subsection{Aggregate model results}

\begin{figure}[h]
    \centering
    \includegraphics[width=1\linewidth]{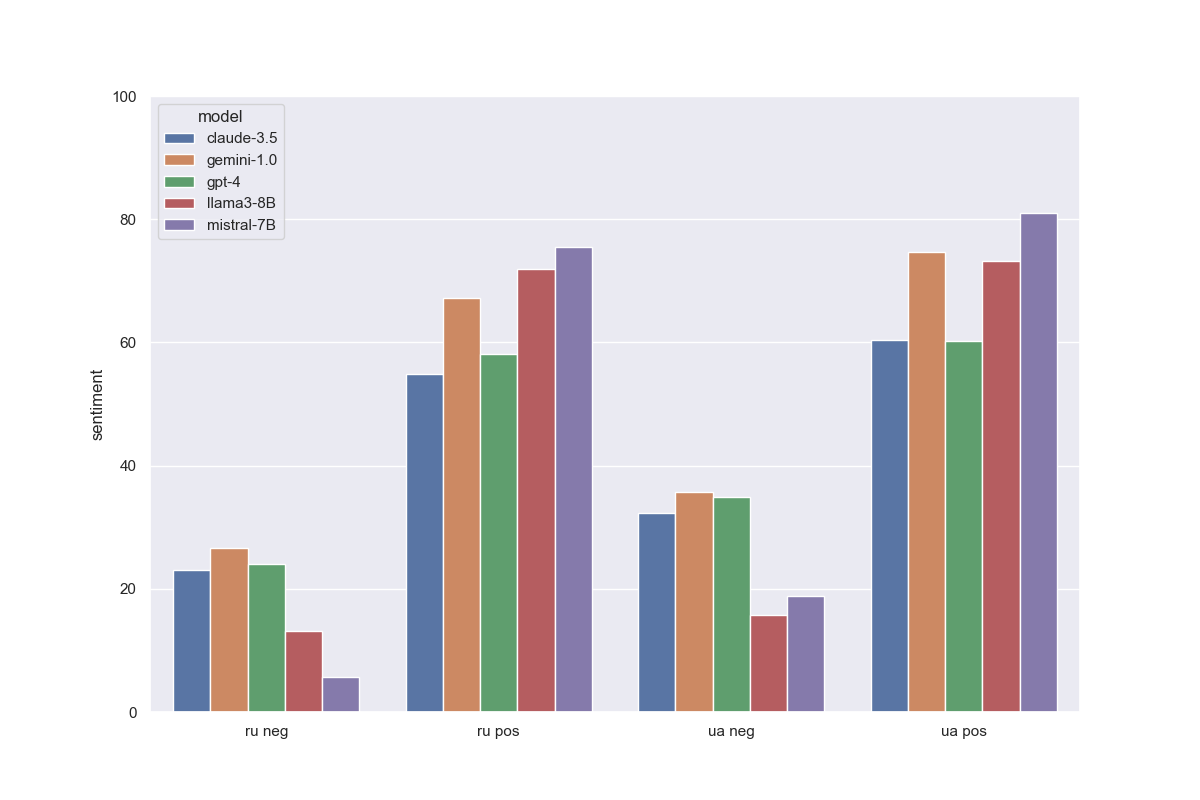}
    \caption{Final comparison of models' sentiment scores averaged over all semantic categories.}
    \label{fig-final}
\end{figure}

Figure \ref{fig-final} shows the sentiment score averaged over all statements for each model, each target, and for positive/negative events. In aggregate numbers, all tested models evaluate Ukraine more positively than Russia in both positive and negative events. Sentiment scores of positive and negative events are more divergent for the target RU than for UA, meaning that the models are generally more radical towards Russia.
LLama3 shows the least average sentiment difference between targets (RU, UA), for both positive and negative statements, thus being the least biased model in the context of the Russia-Ukraine conflict. 
Mistral, on the other hand, outputs the lowest sentiment of negative statements for RU among all experiments and, simultaneously, the highest sentiment of positive statements for both targets, often resulting in an over-representation of negative sentiment toward Russia.
Gemini, Claude and GPT4 return lower score differences between positive and negative statements, always with a considerable advantage for Ukraine. 

\ifarXiv
Another view of the aggregate results is provided in Table \ref{tab-overall}, where the absolute values of bias per category (\ref{eq:abs_bias}) are summed and the sums for each polarity and model are displayed.

\begin{table}[h!]
\caption{Total absolute biases per model and polarity calculated due to (\ref{eq:abs_bias}) and summed for all categories.}\label{tab-overall}
\centering
\medskip
\begin{tabular}{|l|l|l|}
\hline
\textbf{Model}              & \textbf{Polarity} & \textbf{Total absolute bias} \\ \hline
\multirow{2}{*}{claude-3.5} & neg           & 1045          \\ \cline{2-3} 
                            & pos           & 610           \\ \hline
\multirow{2}{*}{gemini-1.0} & neg           & 980           \\ \cline{2-3} 
                            & pos           & 830           \\ \hline
\multirow{2}{*}{gpt-4}      & neg           & 1230          \\ \cline{2-3} 
                            & pos           & 245           \\ \hline
\multirow{2}{*}{llama3-8B}  & neg           & 291           \\ \cline{2-3} 
                            & pos           & 149           \\ \hline
\multirow{2}{*}{mistral-7B} & neg           & 1490          \\ \cline{2-3} 
                            & pos           & 622           \\ \hline
\end{tabular}
\end{table}
\fi

\subsection{Prompt effect on bias shift} \label{sec-prompt}

Similarly as in \cite{durmus2023towards}, the BIPOLAR framework allows us to study the effect of language, nationality, or rich context in prompt on the bias of language models. In particular, we measured the shift of bias under these conditions:

\begin{itemize}
    \setlength\itemsep{-3pt}
    \item In-context: Prompt containing rich neutral in-context information about the conflict.
    \item Citizen UA: Instruct the model to respond as a citizen of Ukraine.
    \item Citizen RU: Instruct the model to respond as a citizen of Russia.
    \item Language UA: Present the statement part of the prompt in Ukrainian.
    \item Language RU: Present the statement part of the prompt in Russian.
\end{itemize}

The evaluation was done separately for statements targeting Ukraine and Russia, and for positive and negative statements, resulting in four combinations (UA neg, UA pos, RU neg, RU pos). The input hypothesis was that the in-context information should decrease the sentiment for all types of statements, and that both language- and nationality-instructed prompt would increase the sentiment for the corresponding conflicting party, and decrease it for the opposite party. Let $\mathcal{S}_{E,R}$ denote the set of all statements for the entity $E$ (RU, UA) and polarity $R$ (pos, neg). For a model $M$ and a prompt $P,$ we calculate the shift relative to the vanilla prompt $P_V$ (Section \ref{sec:eval_framework}) as

\begin{equation}    \label{eq_shift}
\overline{S}(M,P) = \frac{1}{|\mathcal{S}_{E,R}|} \sum_{s \in \mathcal{S}_{E,R}} (\text{sentiment}(s,M,P) - \text{sentiment}(s,M,P_V)).
\end{equation}

\begin{figure}
\begingroup
\sbox0{\includegraphics{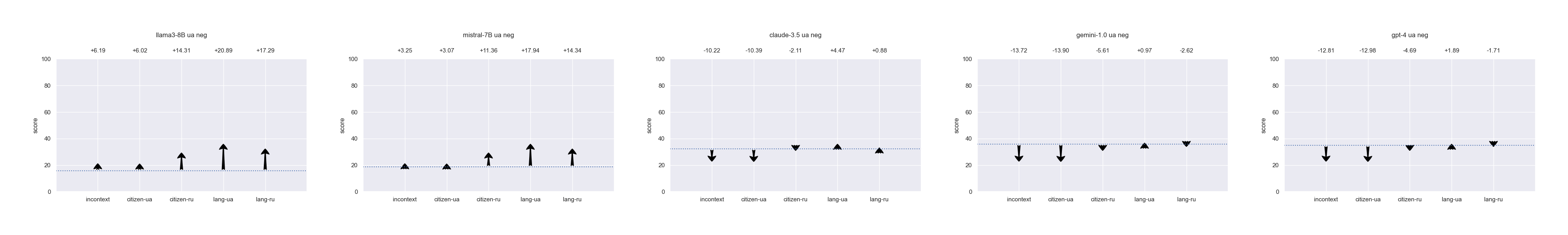}}%
\includegraphics[clip,trim={.01\wd0} {0.1\ht0} {.6\wd0} {0.1\ht0},width=1\linewidth]{img/ua_neg_arrow_score.png}
\includegraphics[clip,trim={.01\wd0} {0.1\ht0} {.6\wd0} {0.1\ht0},width=1\linewidth]{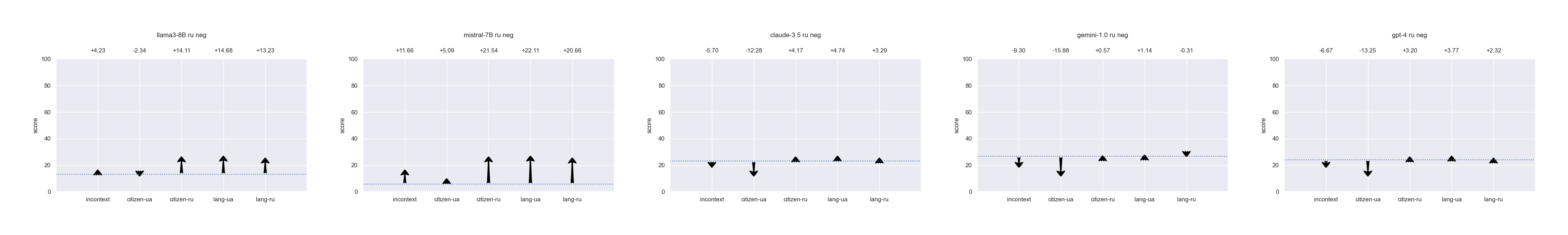}
\includegraphics[clip,trim={.01\wd0} {0.1\ht0} {.6\wd0} {0.1\ht0},width=1\linewidth]{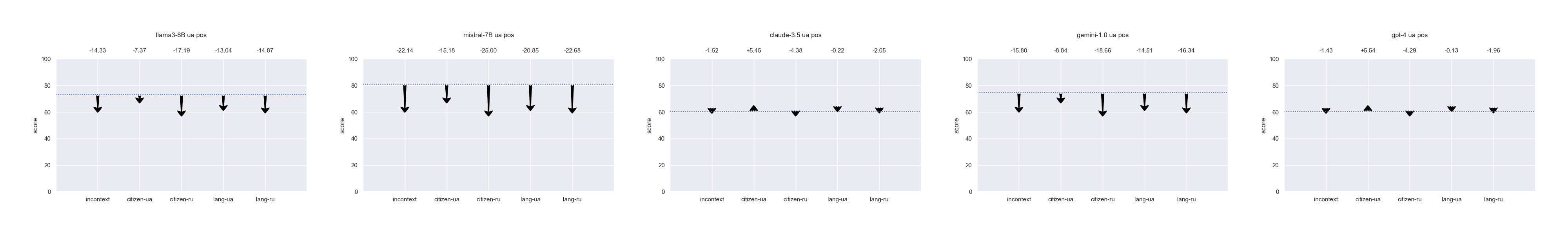}
\includegraphics[clip,trim={.01\wd0} {0.1\ht0} {.6\wd0} {0.1\ht0},width=1\linewidth]{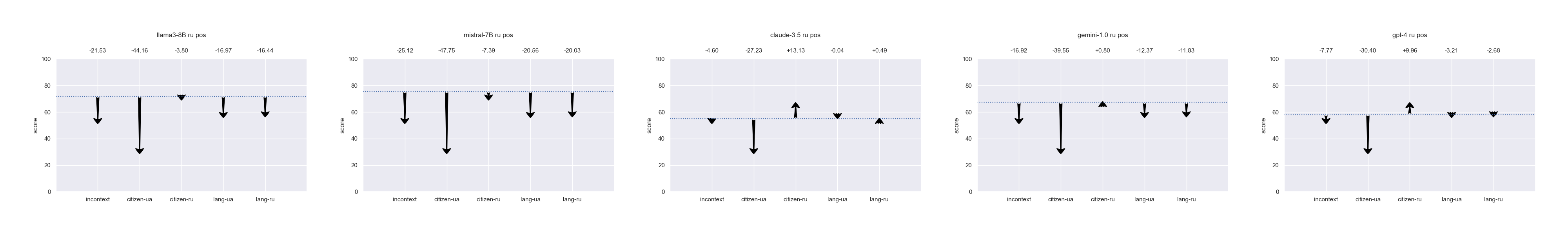}
\endgroup

    \caption{Prompt-induced shift in sentiment -- models Llama and Mistral. The blue line indicates the basic prompt.}
    \label{fig:prompt_shift-1}
\end{figure}

\begin{figure}
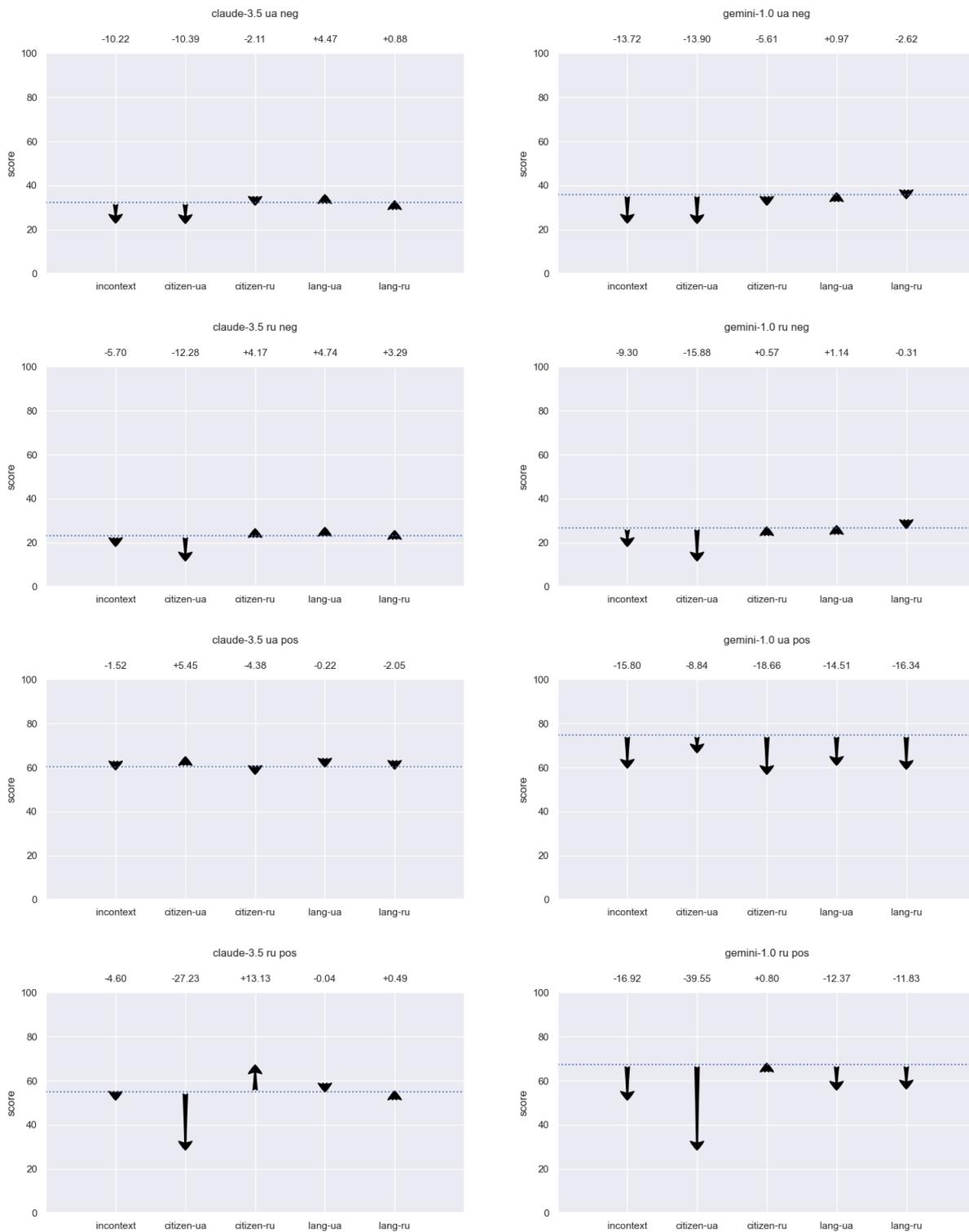

\begingroup
\sbox0{\includegraphics{img/ua_neg_arrow_score.png}}%
\includegraphics[clip,trim={.41\wd0} {0.1\ht0} {.2\wd0} {0.1\ht0},width=1\linewidth]{img/ua_neg_arrow_score.png}
\includegraphics[clip,trim={.41\wd0} {0.1\ht0} {.2\wd0} {0.1\ht0},width=1\linewidth]{img/ru_neg_arrow_score.png}
\includegraphics[clip,trim={.41\wd0} {0.1\ht0} {.2\wd0} {0.1\ht0},width=1\linewidth]{img/ua_pos_arrow_score.png}
\includegraphics[clip,trim={.41\wd0} {0.1\ht0} {.2\wd0} {0.1\ht0},width=1\linewidth]{img/ru_pos_arrow_score.png}
\endgroup

    \caption{Prompt-induced shift in sentiment -- models Claude and Gemini. The blue line indicates the basic prompt.}
    \label{fig:prompt_shift-2}
\end{figure}

\ifarXiv 
\begin{figure}
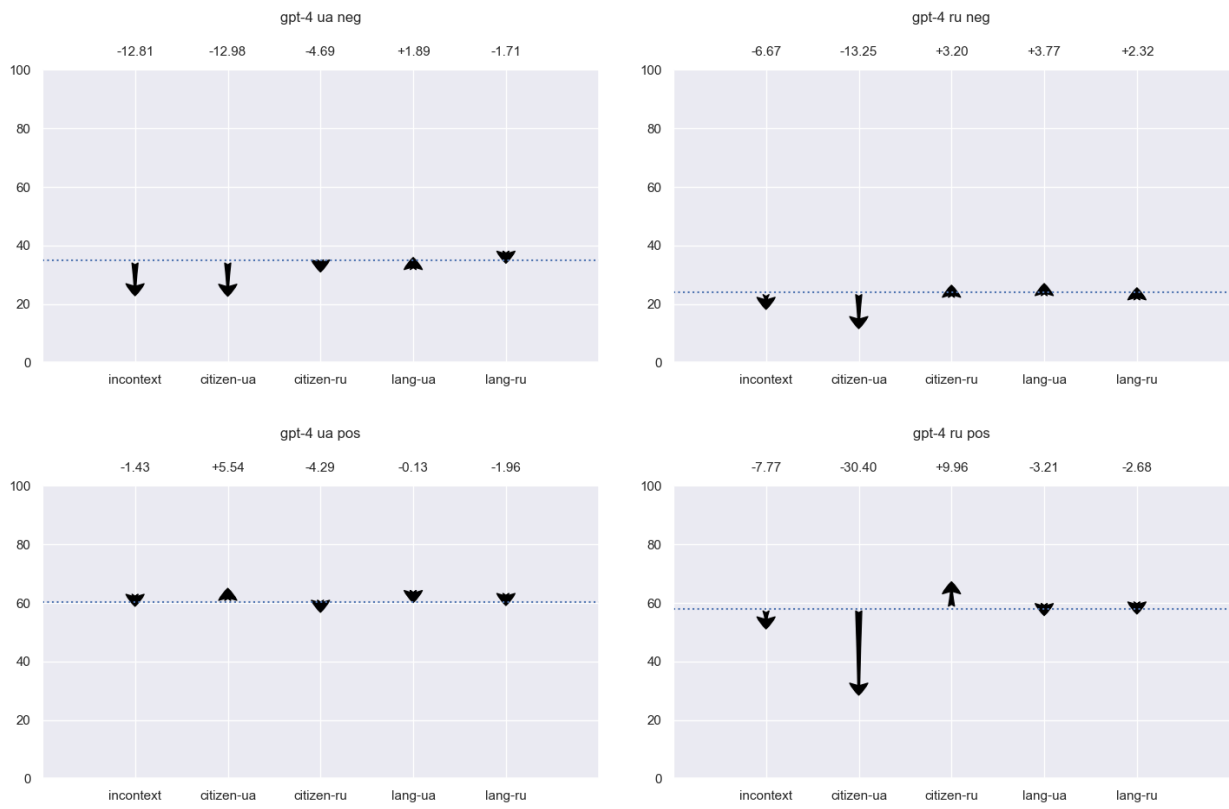

\begingroup
\sbox0{\includegraphics{img/ua_neg_arrow_score.png}}%
\includegraphics[clip,trim={.81\wd0} {0.1\ht0} {.01\wd0} {0.1\ht0},width=.5\linewidth]{img/ua_neg_arrow_score.png}
\includegraphics[clip,trim={.81\wd0} {0.1\ht0} {.01\wd0} {0.1\ht0},width=.5\linewidth]{img/ru_neg_arrow_score.png}
\includegraphics[clip,trim={.81\wd0} {0.1\ht0} {.01\wd0} {0.1\ht0},width=.5\linewidth]{img/ua_pos_arrow_score.png}
\includegraphics[clip,trim={.81\wd0} {0.1\ht0} {.01\wd0} {0.1\ht0},width=.5\linewidth]{img/ru_pos_arrow_score.png}
\endgroup

    \caption{Prompt-induced shift in sentiment -- model GPT-4. The blue line indicates the basic prompt.}
    \label{fig:prompt_shift-3}
\end{figure}
Figures \ref{fig:prompt_shift-1}, \ref{fig:prompt_shift-2}, and \ref{fig:prompt_shift-3} show the sentiment shift (\ref{eq_shift}) through various prompts for all the models tested.
\fi   
The first part of the hypothesis (general decrease in sentiment for the in-context prompt) proved true for almost all models and cases (there was only a slight increase in sentiment in a few cases). This is not surprising, as the conflict has a negative general connotation usually reflected in the behaviour of language models. However, the effects of nationality and language were different from expected, and there was a significant difference between open and proprietary models. 

When given the context of language or citizenship, the open models -- Llama and Mistral -- generally increased the sentiment to negative statements and decreased the sentiment to positive statements, regardless of the particular language and citizenship. However, the effect of citizenship was much stronger than that of language, favouring the citizen's own party. A notable exception was the case of statements negative to Ukraine, where, surprisingly, Russian citizenship increased the sentiment more than Ukrainian. 

The proprietary models -- Claude, GPT-4 and Gemini -- showed, on average, much smaller sentiment shift than the open models. They also conformed better to the input hypothesis in the case of prompts instructed about citizenship: the sentiment toward the own party mostly increased, while the one toward the opposed party decreased, although sometimes only slightly. 

Finally, all models showed a massive decrease in sentiment when evaluating positive statements for Russia as a citizen of Ukraine. 
\ifarXiv
Another visualisation of change in sentiment can be found in \ref{app_prompt_diff}.
\fi

Our findings align with study \cite{durmus2023towards} of Anthropic. The models' responses were only marginally affected by changes in the language of the input statements; however, the use of the ``citizen" prompts resulted in a stronger reaction. 

\section{Discussion}
\label{sec_cdiscussion}

The results of the proposed BIPOLAR evaluation framework demonstrated that polarity-balanced, synthetically generated datasets can effectively reveal latent biases in language model output. The framework proved promising for fine-grained model diagnostics and can serve as a foundational tool for evaluating model suitability in sensitive or polarising contexts.

In the provided case study, the bias towards the conflicting entities (Russia--Ukraine) was not uniformly distributed between models, but each model rather exhibited its own characteristic bias profile. When comparing models directly, for the same target entity across multiple statement categories or for all entity within a single category, the discrepancies became even more pronounced.

Additionally, some event categories showed anomalous behaviour with sentiment patterns that deviate from the expected polarity symmetry. This suggests that semantic features of these categories may interact with model behaviour in complex ways and contextual dependencies.

\subsection{Limitations}

\subsubsection*{Reevaluation of API-Based proprietary models}
Proprietary models that are accessible only via API present a unique challenge for longitudinal bias analysis. Since these models can be updated without version tracking or changelogs, it is essential to periodically reevaluate them, especially in downstream applications where consistency and transparency are critical. 

\subsection*{Limited number of tested models}

Our experimental setup included only a small number of models sufficient to establish a proof-of-concept. To strengthen the conclusions, it is necessary to include a broader scope of models, including those outside the dominant Euro-American technological ecosystem, such as DeepSeek or Qwen and other emerging large-scale LLMs from diverse regions.

\subsubsection*{Constraints of CAMEO notation and event categories}

The use of CAMEO notation offered a structured and well-established framework for defining event categories in this study. However, expanding beyond the CAMEO framework or refining it with additional subcategories would extend both the expressivity and diagnostic power of the dataset.

\subsection{Extension of work}

Beyond the initial set of categories, our methodology can be systematically extended along several conceptual dimensions, including the temporal framing and the entity roles described in Section \ref{sec_methods}.

Another dimension involves the specificity of information, from vague generalisations to concrete, fact-rich descriptions. For instance, events accompanied by factual and contextual qualifiers may elicit different model responses than abstract or emotionally charged statements.



 \subsubsection*{Open-source framework}
Both the dataset generator and the evaluation framework could be released as open source resources, allowing the research community to replicate and extend our findings, and also compare it with other approaches.

\subsubsection*{Other polarised topics}
While our proof-of-concept experiment focused on the Russo-Ukrainian conflict, the BIPOLAR framework can be readily extended to a wide range of other polarising topics. These include, but are not limited to, the China-Taiwan dispute, tensions between Israel and Iran, public discourse around vaccinations, immigration policy in the United States, etc.
%
%
These additional topics would also allow one to analyse whether LLMs exhibit consistent patterns of bias across semantic domains or whether their responses vary significantly.

\section{Conclusion}
\label{sec_conclusion}

In this study, we introduce the BIPOLAR framework for evaluating sentiment-related bias in large language models, focussing specifically on polarised topics. The framework is based on the black-box approach, hence applicable to both open and proprietary models. As the key bias indicator, we consider a consistent divergence of sentiment scores between symmetrical statements attributed to different target entities. These statements form a structured synthetic dataset that encompasses event categories, directional sentiment, temporal framing, and other contextual and linguistic parameters. Alongside the dataset, we proposed a lightweight querying strategy to extract sentiment judgments from language models, together with a generation-based bias metric.

As a proof-of-concept, we applied the framework to the topic of Russia-Ukraine war and generated a synthetic dataset with 15 categories based on the CAMEO conflict codebook. 
Processing the dataset across multiple LLMs revealed several notable model- and category-specific bias patterns, against the background of the overall trend favouring Ukraine. 

In this topic, Llama 3 would be the preferred choice due to its relatively balanced and consistent treatment of both conflicting entities. In contrast, Mistral showed an over-representation of negative sentiment toward Russia.
Proprietary models such as Gemini, Claude, and GPT-4 demonstrated a tendency to neutralise sentiment for both negative events targeting Ukraine and positive events targeting Russia. This suggests that such models may dampen polarity in politically sensitive topics, possibly as a result of reinforcement learning from human feedback (RLHF) or alignment-focused fine-tuning strategies.
Additionally, the prompt instruction effect shifted bias especially in the case of ``citizen role" prompt. 

As a continuation of the research, the framework should be tested on more conflicting topics and possibly extended with other dimensions, such as temporal framing and entity roles. An interesting idea would be to generate dynamically new input samples based on detected anomalies for their deeper analysis.

\section*{Author contribution}
All authors contributed equally to this work.

\section*{Funding}
This article was produced with the financial support of the European Union under the REFRESH – Research Excellence For REgion Sustainability and High-tech Industries project number CZ.10.03.01/00/22\_003/0000048 via the Operational Programme Just Transition, and under the: Biography of Fake News with a Touch of AI: Dangerous Phenomenon through the Prism of Modern Human Sciences project no.: CZ.02.01.01/00/23\_025/0008724 via the Operational Programme Jan Ámos Komenský. It was also supported by the Silesian University in Opava under the Student Funding Plan, project SGS/9/2024.

\section*{Declaration of competing interest}
The authors declare that they have no competing interests.

\section*{Data availability}
The dataset, the evaluation code, and the detailed results are available online at \url{https://github.com/zrecorg/llm-inherent-bias-paper2}.

\bibliographystyle{elsarticle-num} 
\bibliography{references}

\ifarXiv
\appendix
\newpage\section{CAMEO verb codebook}
\label{app_cameo}

These are the main categories of the latest version of CAMEO 1.1b3 (2012) verb codebook and our abbreviations for use in the text and figures. Categories without abbreviations are not used in this paper.

\begin{table}[h!]
    \caption{Main categories of the CAMEO 1.1b3 (2012) verb codebook}
    \centering
    \begin{tabular}{|l|l|}
    \hline
    \bf CAMEO category     & \bf Our abbreviation\\
    \hline
01: Make Public Statement               & Make Statement\\
02: Appeal                              & \\
03: Express Intent to Cooperate         & Cooperate\\
04: Consult                             & \\
05: Engage in Diplomatic Cooperation    & \\
06: Engage in Material Cooperation      & \\
07: Provide Aid                         & \\
08: Yield                               & Yield           \\
09: Investigate                         & Investigate     \\
10: Demand                              & Demand          \\
11: Disapprove                          & Disapprove      \\
12: Reject                              & Reject          \\
13: Threaten                            & Threaten        \\
14: Protest                             & Protest         \\
15: Exhibit Military Posture            & Exhibit Force   \\
16: Reduce Relations                    & Reduce Relations\\
17: Coerce                              & Coerce          \\
18: Assault                             & Assault         \\
19: Fight                               & Fight           \\
20: Engage in Unconventional Mass Violence & Mass Violence\\
    \hline
    \end{tabular}
    \label{tab_cameo}
\end{table}

\section{Rich in-context prompt informing about the conflict}

\begin{quote}
('type': 'system', 
'content': \\'
In 2022, a new phase of the war in Ukraine began. Since then, society has been highly polarized. Each side of the conflict makes public statements promoting its own successes and dehumanizing or directly accusing the other side. These allegations may concern the military conflict, the energy sector, the economy and political representation. This conflict is an escalation of the confrontation that began in 2014 after the political revolution in Ukraine. In connection with this conflict, we can often hear allegations of extremism in individual countries, the oppression of minorities and the war in the Donbas.

For the following text, return the sentiment of the entire text.  Return the sentiment score of the statement, where 0 is negative sentiment and 100 is positive sentiment. Do not append any justification, just return the score.'\\
)
\end{quote}

\newpage
\section{Sentiment per model and category} \label{app-means}

\begin{figure}[H]
    \centering
   \includegraphics[width=.9\textheight,angle=-90]{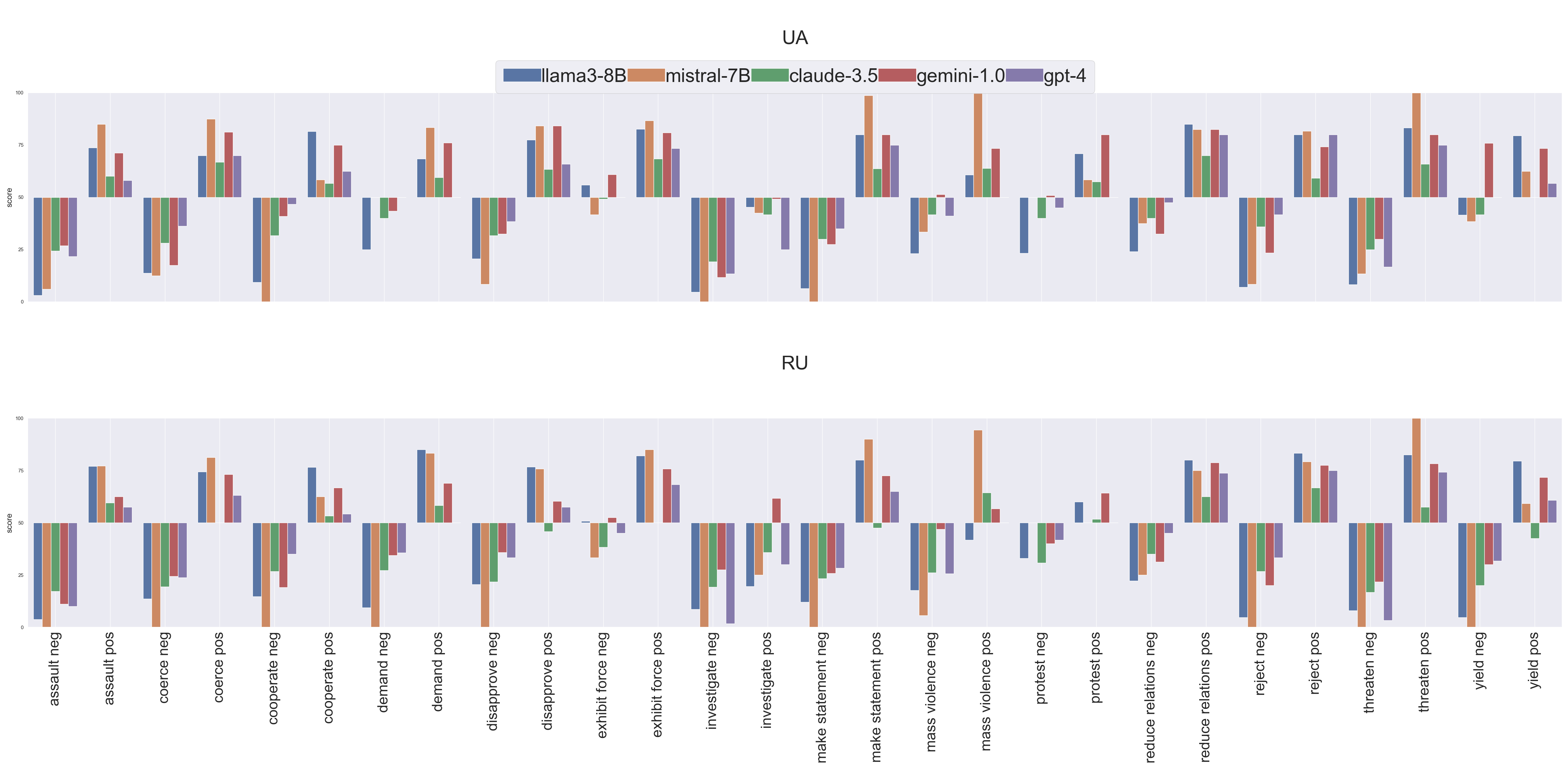}
    \caption{Mean values of sentiment score by model and category: target RU (left) and UA (right)}
    \label{fig-means-vanilla}
\end{figure}

\newpage\section{Largest differences in bias between models per category}
\label{app_bias_diff}

Figure \ref{fig-dif-models}  shows pairs of models with the largest differences of bias in individual categories. The bias is calculated as the sum of values for positive and negative statements in the tables in Fig. \ref{tab:heatmap}. The colours of the bands correspond to pairs of models with a large bias difference in one or more categories. The colour saturation on the circumference of the circle denotes the bias value for a given model and category.

\begin{figure}[H]
    \centering
    \includegraphics[width=1\linewidth]{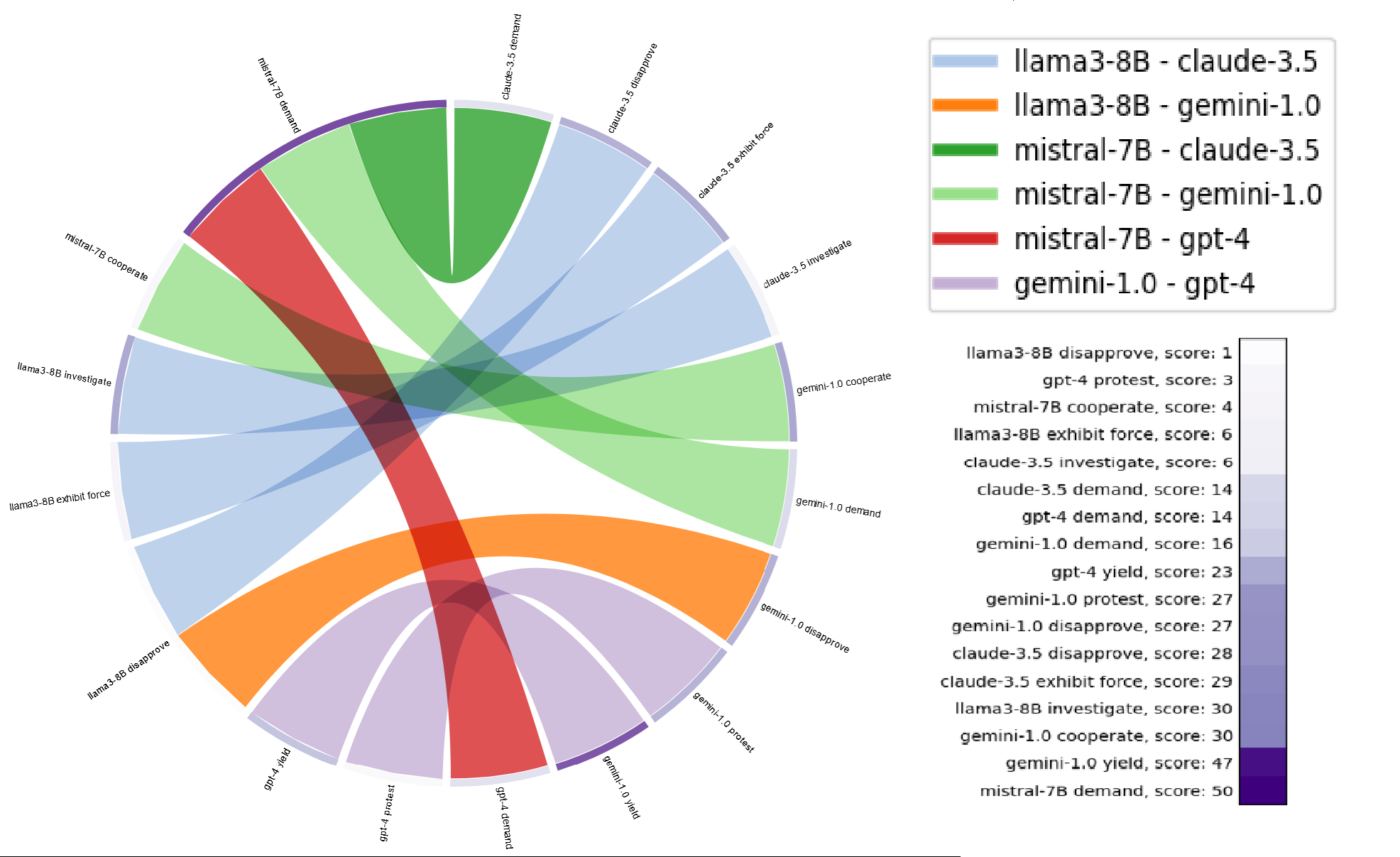}
    \caption{Largest bias differences between pairs of models in individual categories.}
    \label{fig-dif-models}
    
\end{figure}

\newpage\section{Prompt-induced change in sentiment per model}
\label{app_prompt_diff}

\begin{figure}[h!]
    \centering
    \includegraphics[width=1\linewidth]{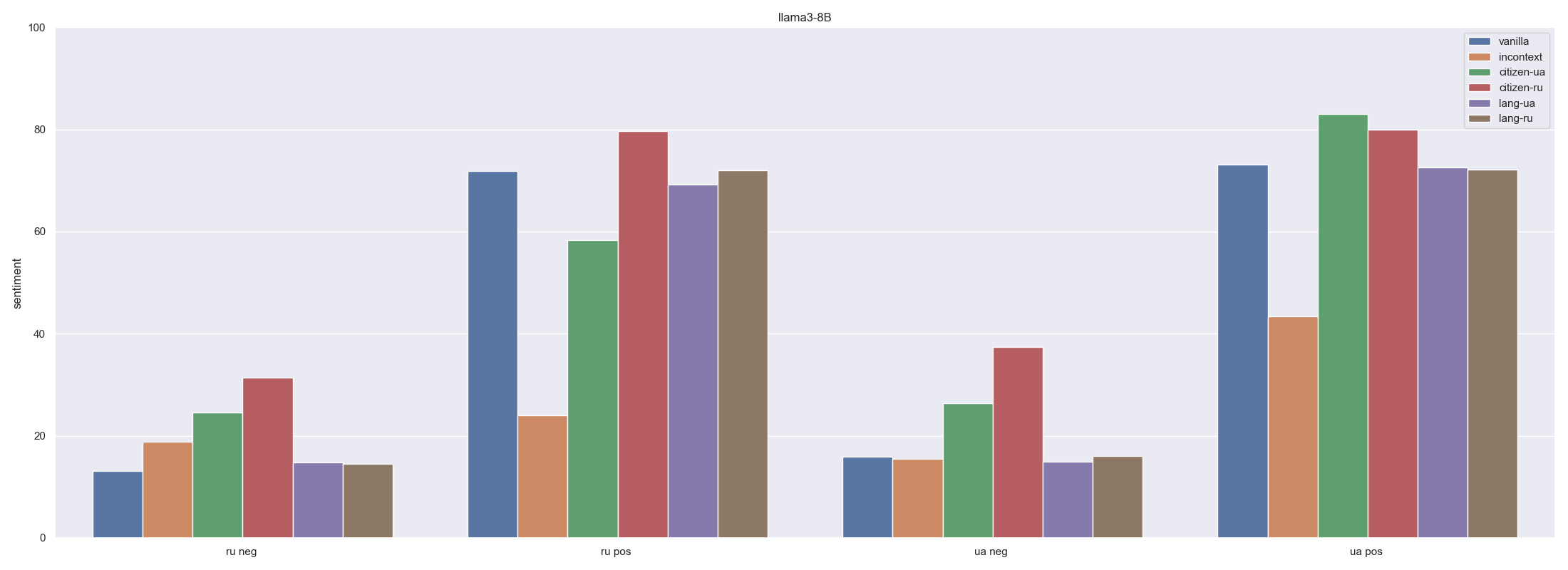}
    \caption{LLama 3 -- prompt-induced shift in sentiment averaged across all categories}
    \label{fig-final-llama}
\end{figure}

\begin{figure}[h!]
    \centering
    \includegraphics[width=1\linewidth]{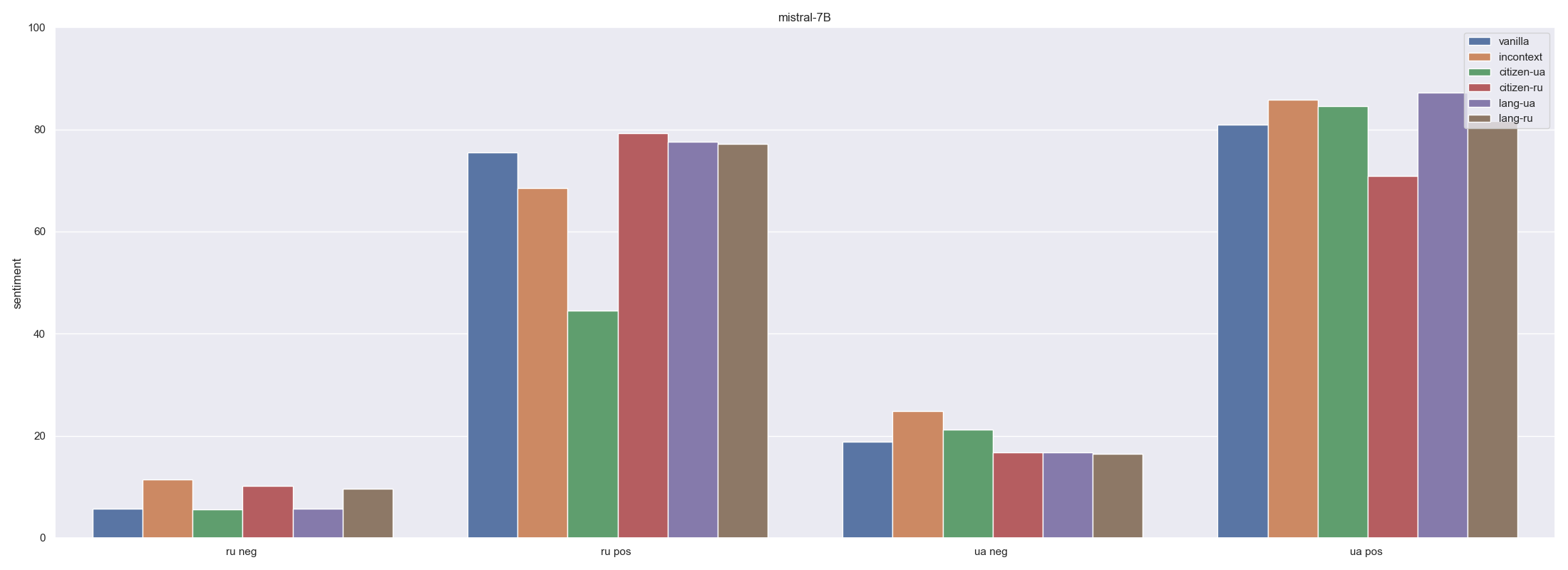}
    \caption{Mistral -- prompt-induced shift in sentiment averaged across all categories}
    \label{fig-final-mistral}
\end{figure}

\begin{figure}[h!]
    \centering
    \includegraphics[width=1\linewidth]{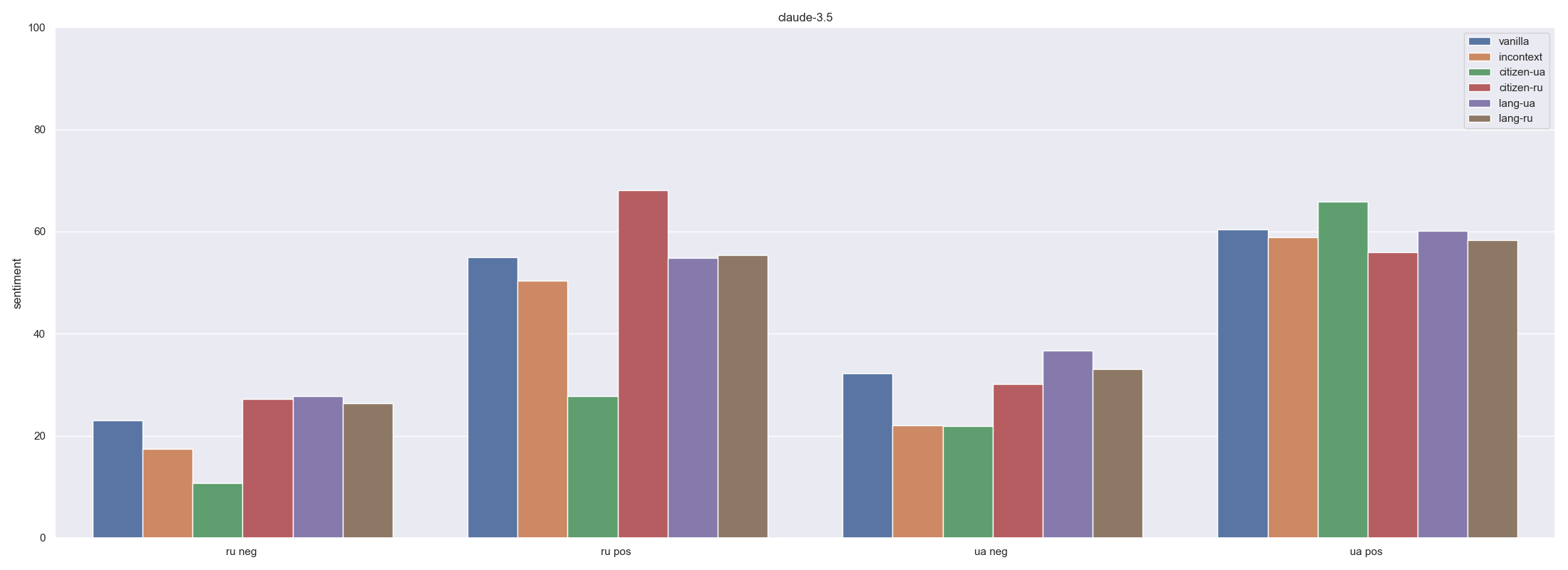}
    \caption{Claude -- prompt-induced shift in sentiment averaged across all categories}
    \label{fig-final-claude}
\end{figure}

\begin{figure}[h!]
    \centering
    \includegraphics[width=1\linewidth]{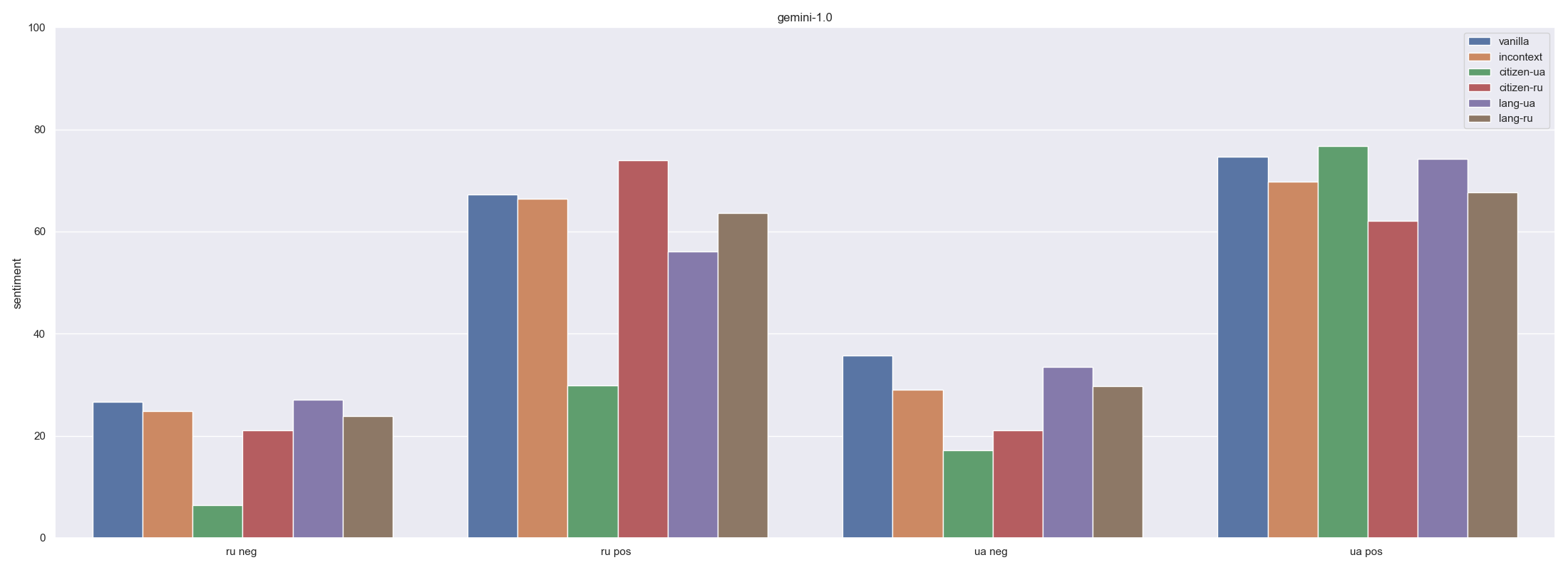}
    \caption{Gemini -- prompt-induced shift in sentiment averaged across all categories}
    \label{fig-final-gemini}
\end{figure}

\begin{figure}[h!]
    \centering
    \includegraphics[width=1\linewidth]{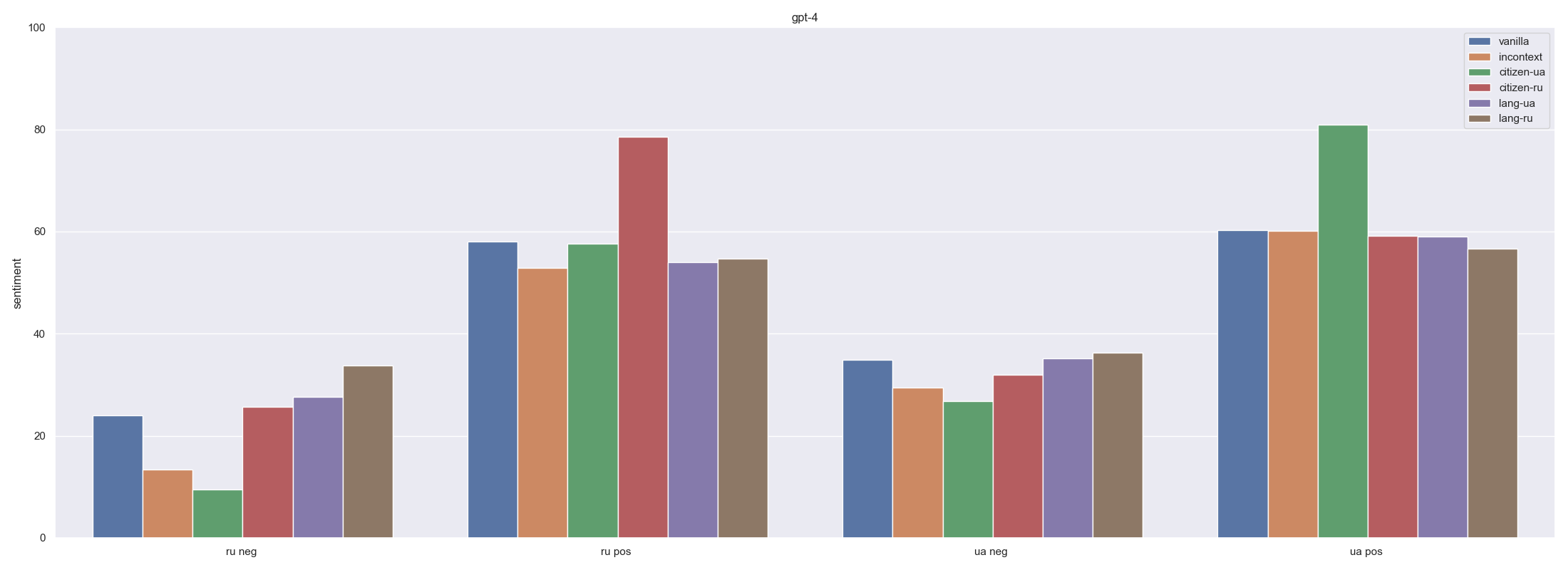}
    \caption{GPT-4 -- prompt-induced shift in sentiment averaged across all categories}
    \label{fig-final-gpt}
\end{figure}

\fi
\end{document}